\documentclass{article}

\usepackage{booktabs}
\usepackage{graphicx}
\usepackage{url}
\usepackage{amsmath,amssymb}
\usepackage{caption}
\usepackage{subcaption}
\usepackage{hyperref}
\usepackage{xcolor}
\hypersetup{
    colorlinks,
    linkcolor={red!50!black},
    citecolor={blue!50!black},
    urlcolor={blue!80!black}
}
\usepackage{comment}
\usepackage{amsmath,amssymb} 
\usepackage{tabularx}
\usepackage{multirow}
\usepackage{caption}
\usepackage{adjustbox}
\usepackage{algorithm}
\usepackage{algorithmic}

\usepackage[top=1in, bottom=1.25in, left=1.25in, right=1.25in]{geometry}

\title{\textbf{Improving saliency models' predictions of the next fixation with humans' intrinsic cost of gaze shifts}}
\author{Florian Kadner$^{*}$ \quad Tobias Thomas$^{*}$ \quad David Hoppe$^{*}$ \quad Constantin A. Rothkopf}
\date{Centre for Cognitive Science \& Institute of Psychology \\ Technical University Darmstadt, Germany \\ \vspace{0.1cm}
$^{*}$The authors contributed equally to this work}
\begin{document}
\maketitle
\begin{abstract}
The human prioritization of image regions can be modeled in a time invariant fashion with saliency maps or sequentially with scanpath models. However, while both types of models have steadily improved on several benchmarks and datasets, there is still a considerable gap in predicting human gaze. Here, we leverage two recent developments to reduce this gap: theoretical analyses establishing a principled framework for predicting the next gaze target and the empirical measurement of the human cost for gaze switches independently of image content. We introduce an algorithm in the framework of sequential decision making, which converts any static saliency map into a sequence of dynamic history-dependent value maps, which are recomputed after each gaze shift. These maps are based on 1) a saliency map provided by an arbitrary saliency model, 2) the recently measured human cost function quantifying preferences in magnitude and direction of eye movements, and 3) a sequential exploration bonus, which changes with each subsequent gaze shift. The parameters of the spatial extent and temporal decay of this exploration bonus are estimated from human gaze data. The relative contributions of these three components were optimized on the MIT1003 dataset for the NSS score and are sufficient to significantly outperform predictions of the next gaze target on NSS and AUC scores for five state of the art saliency models on three image data sets. 
Thus, we  provide an implementation of human gaze preferences, which can be used to improve arbitrary saliency models' predictions of humans' next gaze targets.

\textcopyright 2023 IEEE. Published in: \url{https://doi.org/10.1109/WACV56688.2023.00214}

\end{abstract}

\maketitle

\section{Introduction}
Because of the inhomogeneous spatial acuity of the visual system, humans shift their gaze sequentially across visual scenes using saccadic gaze movements \cite{findlay2003active}. Four main factors have been shown to influence observers' eye movements: the ongoing task, image features such as contrast and intensity, semantic features such as faces and scene context, but also factors that arise from the sequential interaction of an observer with the scene including center bias, proximity preference, inhibition of return \cite{tatler2011eye}. Empirically, gaze targets of multiple human observers while inspecting an image given different task instructions can be collected.
Assuming that gaze prioritization is image-computable, 
the task of predicting 
gaze prioritization given an image
is referred to as visual saliency modeling resulting in a time invariant saliency map, whereas scanpath models generate a sequence of 
gaze targets.

While originally developed to account for the phenomenon of pop-out \cite{treisman1980feature}, visual saliency modeling has been generalized to predicting 
the likelihood of human observers 
overtly shifting their gaze to image regions for arbitrary images.
Initially, saliency models used handcrafted features inspired by neurophysiological properties of the visual system \cite{itti_model_1998,xiaoshuai_sun_what_2012}, but more recently data driven approaches \cite{huang_salicon_2015,kruthiventi_deepfix_2015,pan_shallow_2016,kummerer_understanding_2017-1,cornia_predicting_2018,wang_revisiting_2018,CASNet,EML_Net} have commonly used features from DNNs pretrained on large image datasets, thereby improving performance on various benchmarks \cite{borji2012state,borji_saliency_2019}.
These improvements are due to the rich image structure learned by DNNs when trained on large image datasets, e.g. the VGG19 \cite{simonyan_very_2015} underlying Deep Gaze II \cite{kummerer_understanding_2017-1} is trained on object recognition of one Million images before tuning to saliency problems using the SALICON dataset \cite{jiang_salicon_2015} containing 10000 images.

Scanpath models, by contrast, take an image as input and generate a full scanpath, i.e. a sequence of individual fixation locations as output \cite{wang_simulating_2011,boccignone_modelling_2004,liu_semantically-based_2013,xia2019predicting,assens_saltinet_2017,leal-taixe_pathgan_2019}.
Progress on a variety of benchmarks and image databases has been made, but, a fundamental difficulty with scanpath models compared to saliency models is the well known variability of individual gaze sequences between observers, which poses particular challenges for evaluating the quality of predictions.
A recent comprehensive theoretical and empirical evaluation and comparison of scanpath models \cite{kummerer2021state} has revealed that scanpath similarity metrics can score wrong models better than the generating model.
The resulting analysis in \cite{kummerer2021state} establishes that 
a more consistent and meaningful task consists in 
the prediction of the next fixation target conditional on past fixations within an image, which is the task we adopt in this study.

Here, we leverage two recent developments to improve the prediction of the next fixation of human observers given an arbitrary saliency map and the sequence of preceding fixations: the theoretical analysis of scanpath models \cite{kummerer2021state} and the recently measured human cost function for gaze shifts \cite{hoppe2022costs}.
We adopt a computational account of the scanpath as a sequential decision process in the spirit of previous approaches \cite{jiang_learning_2016,mathe2013action,hoppe2019multi}, but differently from these approaches, our algorithm can utilize arbitrary saliency maps as input instead of estimating rewards for image features from scratch. 
First, we reason that saliency corresponds to the reward associated with the free-viewing task which is approximated by marginalizing over all visual tasks. The reason is, that the free-viewing task is maximally ambiguous regarding its task goal.
Second, 
our formulation allows incorporating a map representing the human preferences for gaze shifts, which have recently been estimated for the first time independently of image content through a human psychophysical experiment \cite{hoppe2022costs}. This gives a computational explanation for commonly used heuristics including the proximity preference.
Third, we account for past fixations through a temporally changing exploration map and present the resulting predictions of subsequent gaze targets. The relative contributions of these three components were optimized on the MIT1003 dataset for the NSS score and are sufficient to significantly outperform predictions of the next gaze target on NSS and AUC scores for five state of the art saliency models on three image data sets.  

\section{Related Work}
The concept of saliency lies at the intersection of cognitive science, neuroscience, and computer vision \cite{tatler2011eye}. Empirically, human gaze targets depend strongly on the ongoing task \cite{hayhoe2005eye} but humans tend to look preferentially at certain areas even when free-viewing images \cite{henderson1998eye}. These observations have been complemented with the discoveries of multiple retinotopic maps in the visual system \cite{treue2003visual}. 
While the exact relationships between attention, gaze sequences, and their neuronal underpinnings are still heavily debated, 
visual saliency modeling has become a canonical computer vision task.

Initially, saliency models used handcrafted lower level features like intensity, color, and orientation \cite{itti_model_1998}, whereas current DNN based algorithms determine salient regions in a data driven fashion by reusing learnt features e.g from CNNs \cite{kruthiventi_deepfix_2015}. 
Other studies have emphasized the importance of higher level information in images, such as text and faces \cite{cerf2008predicting,bylinskii_what_2019} or general semantic content \cite{henderson_meaning_2019,pedziwiatr_meaning_2019}. Relevance might be biased, e.g. towards text, see \cite{leal-taixe_pathgan_2019} for a discussion. 
Some approaches have incorporated task goals into models of gaze selection \cite{navalpakkam2005modeling,borji2012probabilistic}, albeit with a small number of tasks with respect to the broad range of human visual and visuomotor tasks.
Semantic information has been incorporated by neural network approaches, for example by pretraining on object recognition \cite{kummerer_understanding_2017-1,huang_salicon_2015,pan_shallow_2016,cornia_predicting_2018,kruthiventi_deepfix_2015,wang_revisiting_2018}. The work on attention in DNNs, e.g \cite{mnih_recurrent_nodate,welleck_saliency-based_nodate} is somewhat complimentary \cite{adebayo2018sanity}, as it is not necessarily modeling overt shifts of attention by gaze shifts but sequential processing of internal representations. Overall, visual saliency modeling is an established field with canonical datasets and benchmarks and
progress on these benchmarks has been steady \cite{borji2012state,borji_saliency_2019}.

Scanpath models have received less attention compared to saliency models but recent approaches include models based on biological and cognitive facts \cite{itti_model_1998,wang_simulating_2011,zanca2019gravitational}, statistically motivated models \cite{boccignone_modelling_2004,liu_semantically-based_2013,xia2019predicting}, and models, which leverage machine learning techniques for prediction without reference to underlying mechanisms of vision \cite{assens_saltinet_2017,leal-taixe_pathgan_2019}. 
While some algorithms require an image as input \cite{itti_model_1998,kummerer_understanding_2017-1,xiaoshuai_sun_what_2012,CASNet,EML_Net,SAM_Resnet,drostejiao2020},
other models use a saliency map as input for generating a scanpath \cite{itti_model_1998,wang_simulating_2011,zanca2019gravitational,assens_saltinet_2017,leal-taixe_pathgan_2019,boccignone_how_nodate,boccignone_modelling_2004,xiaoshuai_sun_what_2012,xia2019predicting}.
In \cite{boccignone_modelling_2004,boccignone_how_nodate} the authors investigated the properties of scanpaths as function of parameters in random walks on saliency maps. \cite{wang_simulating_2011} proposed a model incorporating an image representation map based on filter responses, foveation, and a memory module to generate sequential saliency maps. \cite{xiaoshuai_sun_what_2012} used an algorithm based on projection pursuit to select image targets for simulating scanpath in order to mimic the sparsity of human gaze selection. PathGAN \cite{leal-taixe_pathgan_2019} extracts image features using available DNNs and trains recurrent layers to generate and discriminate scanpaths in a training set. While PathGAN learned scanpaths end-to-end and outperformed several other models, qualitative results suggest persistent deviation to scanpaths of human observers.

Of particular relevance in this context is recent work on characterizing and evaluating the prediction accuracy of scanpath models relative to human gaze \cite{kummerer2021state}.
The authors' in depth analyses show that 
some scapath similarity metrics such as ScanMatch \cite{cristino2010scanmatch} or MultiMatch \cite{jarodzka2010vector} can score wrong models better than the generating model given ground truth.
Note also, that some of the current scanpath models employ statistics of scanpaths as a means to capture behavioral biases of gaze shifts, but these have so far never been measured independently of image content.
The in depth analyses in \cite{kummerer2021state} convincingly lead to the conclusion, that instead of comparing entire scanpaths it is more adequate to evaluate models regarding their prediction of the next fixation within a given scanpath.

Finally, scanpaths have also been conceptualized as sequential decision problems, which is particularly successful in situations where
observers' goals are known and can therefore be
formalized as rewards 
\cite{mnih_recurrent_nodate,hoppe2016learning,hoppe2019multi}. Very much related to the present approach, \cite{mathe2013action} used inverse RL to estimate implicit rewards from human gaze sequences. While this extracts reward functions in terms of image features, it is agnostic in relation to internal, behavioral costs and benefits.
Other studies have used reinforcement learning to predict scanpaths \cite{jiang_learning_2016} with a state consisting of low-level features, semantic features, center bias, spatial distribution of eye-fixation shifts as well as a measure indicating previous gaze visits. However, these studies did not
utilize the human cost for eye movements measured independently of image content and predicted 'fixation stages' in an experiment and not individual fixations. But empirical studies have shown, that oculomotor biases are not independent of image content, e.g. by simply rotating images \cite{foulsham2008turning}.
Here, our goal is to leverage the recently measured human cost for making a gaze shift independently of image content \cite{hoppe2022costs} for arbitrary saliency models so that we do not infer the rewards of image features from scratch. This allows arbitrary saliency models to improve their predictions of the next gaze target by incorporating the intrinsic costs of a gaze shift in human observers, which interact with the prioritization of image content. 



\section{One-step ahead prediction model}
Our general model is based on statistical decision theory and gaze sequences are viewed as reward-driven behavioral sequences, that can be described using a Markov Decision Process (MDP), similar to \cite{mathe2013action,jiang_learning_2016,hoppe2019multi}.
A scanpath is a sequence of gaze locations $\mathbf{x_0}, \mathbf{x_1}, \dots, \mathbf{x_t}$ visited on an Image $I$ through movement of the visual apparatus.
Each of the visited gaze locations is the result of a decision for that particular location, following a policy $\pi(s) = \text{arg}\max\limits_\mathbf{x_{t+1}} Q(s,\mathbf{x_{t+1}}) $, where $Q(s,\mathbf{x_{t+1}}) = \mathbb{E}\left[G_t\mid s, \mathbf{x_{t+1}}\right]$ are the Q-values, i.e. the expected discounted total future rewards $G_t=\sum_{i=1}^{N}\gamma^{i-1}r_{t+i}$ when switching gaze to a location $\mathbf{x_{t+1}}$ while being in state $s$.
The state $s$ summarizes relevant factors that contribute to the selection of the next action $\mathbf{x_{t+1}}$, $\gamma$ is the discount factor, and $N$ is the number of gaze shifts until the end of looking at a particular image.
When exploring an image $I$, action selection is affected by past eye movements as well as the image, therefore $s = (I, \mathbf{x_0}, \dots, \mathbf{x_t})$.
For some tasks, $s$ might also include further task-relevant features or it could represent a
belief state.
We argue that this is allowable here because saliency can be thought of as an average over all possible states within all possible tasks as we will formulate in the following.

As has been shown repeatedly in the past, human action selection, in particular the generation of eye movements, is driven by multi-dimensional reward structures.
However, the precise composition of the sources of rewards is usually unknown or not easy to measure.
Here, we consider three components that have been shown to drive action selection: task-related reward, behavioral costs, and 
sequential effects related to the history of previous actions.
In order to compute the state-action values $Q_{\text{task}}(s, \mathbf{x_{t+1}}) = \mathbb{E}\left[G\mid s,\mathbf{x_{t+1}}\right]$ in a specific task, we need to specify the rewards:
\begin{align}
 r(s,\mathbf{x_{t+1}}) =  w_0 r_{\text{task}}(s, \mathbf{x_{t+1}}) + w_1 r_{\text{internal}}(s,\mathbf{x_{t+1}}) + w_2 r_{\text{fixation history}}(s,\mathbf{x_{t+1}})
 \label{eq:rtotal}
\end{align}
where $\mathbf{x_{t+1}}$ is a potential next eye movement location and $r_\text{internal}$, $r_\text{task}$ and $r_\text{fixation history}$ are components contributing to the state-action value.

\subsection{Saliency in the context of rewards}
One dimension contributing to action selection is task-related reward.
Eye movements have been shown to be carried out to lead to high rewards in their respective tasks, such as visual search \cite{najemnik2005optimal}, image classification \cite{peterson2012looking}, and can even be planned \cite{hoppe2019multi}.
For free viewing of natural images, however, the reward function is difficult to obtain theoretically because the task instructions are highly ambiguous: "Just look around.".
Here, we conjecture that saliency can be thought of as an average reward over all possible states within all possible tasks as we will formulate in the following.
One possible approach is to formulate the task-related reward structure of free viewing as the result of marginalizing over all possible tasks:
\begin{align}
r_\text{free view}(s, \mathbf{x_{t+1}}) &= \mathbb{E}_{\text{task}} \; \big{[} \mathbb{E}_{s_\text{task}}  \left[r_{\text{task}}(I, s_\text{task}, \mathbf{x_{t+1}})\right]\big{]} \nonumber \\
 &=  \mathbb{E}_{\text{task}} \; \left[ \int\limits_{s_\text{task}} r_{\text{task}}(I, s_\text{task}, \mathbf{x_{t+1}}) p(s_\text{task}) ds_\text{task} \right] \nonumber \\
 &= \sum_{\text{task}} \; \int\limits_{s_\text{task}} r_{\text{task}}(I, s_\text{task}, \mathbf{x_{t+1}}) p(s_\text{task}) ~ ds_\text{task} ~ p(\text{task}) \nonumber \\
 &\approx S(I,\mathbf{x_{t+1}})
\label{eq:rtask}
\end{align}
where $r_{\text{task}}(I, s_\text{task}, \mathbf{x_{t+1}})$ denotes the reward when performing eye movement $\mathbf{x_{t+1}}$ in image $I$ under a specific task while being in state $s_\text{task}$.
The state $s_\text{task}$ summarizes all relevant information about the actions performed prior to the current decision for a specific task.
The probability distribution over potential tasks $p(\text{task})$ weights the task-dependent reward according to how likely the task is.
For example, information that is relevant for many visual tasks, e.g., faces, receives higher weights.
Finally, $p(s_\text{task})$ is the probability distribution of the current state within a task, i.e. the action sequence (scanpath) prior to the current fixation and $S(I,\mathbf{x})$ is the saliency score.

In conclusion, we view the saliency of an image location during free viewing as the 
approximate average reward of that location across all possible tasks and all possible previous gaze shifts in that task. 

\subsection{Influence of past fixations}
According to Equation \ref{eq:rtask} we can approximate the task related component to the reward using the predictions of a saliency model.
However saliency models are time-invariant, depending only on the image.


Here, we propose an extension to overcome this problem and compute saliency models taking into account past actions. 
Our approach is based on the fact that the next fixation depends on all prior fixations. Since the exact nature of this relationship is unknown, we developed a model that quantifies the influence of a fixation within a gaze sequence on the selection of future fixation choices:
\begin{align}
r_\text{fixation history}(s, \mathbf{x_{t+1}}) &= r(\mathbf{x_{0}},\dots,\mathbf{x_{t-1}},\mathbf{x_t}, \mathbf{x_{t+1}}) = \sum_{i=0}^{t} \phi_i ~ \mathcal{N}(\mathbf{x_{t+1}};\mathbf{x_{i}},\mathbf{\Sigma}) \label{eq:fix_hist}
\end{align}
Positive values ($\phi_i > 0$) indicate that having visited location $\mathbf{x_i}$ at timestep $i$ during the same scanpath increases the probability of targeting the next fixation to location $\mathbf{x_i}$.
Negative values lead to reduced probabilities, therefore corresponding to an effect such as a spatial version of inhibition of return.

This reward can be conceptualized as the trade-off between exploration and exploitation, i.e. a reward for either parts of the state-space that have never been explored, or, if the environment can change over time, have not been explored recently \cite{sutton1990integrated}. Equivalently, this reward can be formulated as an exploration bonus.
Therefore, this part of the reward structure encourages an agent to try long-ignored actions, i.e. visit locations that have not been visited yet or have not been visited in a long time. 
Since the exact nature of this relationship is yet to be understood, we estimated the parameters $\phi_i$ from the eye movement data.
Note that we did not constrain the parameters to sum up to one, to allow both positive and negative values for already visited or not recently visited regions, in principle.

\subsection{Oculomotor preference map}
\label{sec:oculo}
Saliency models commonly neglect the agent's effort expended in the actual action to gain visual information, although such internal costs influence gaze shifts \cite{hoppe2016learning,hoppe2019multi}. These costs and benefits have their origin in the effort to produce the movement, which includes cognitive costs such as deciding upon where to move next \cite{hoppe2019multi} and when \cite{hoppe2016learning}.
The oculomotor preferences were recently measured independently of image content in a psychophysical experiment involving a preference elicitation paradigm \cite{hoppe2022costs}. 
Subjects repeatedly chose between two visual targets by directing gaze to the preferred target.
Eye movements were recorded using eye tracking.
For each choice, three properties were manipulated for both targets: the distance to the current fixation location, the absolute direction to the target (e.g., left), and the angle relative to the last saccade.
Using the decisions we inferred the value of each component and integrated them in an oculo-motor preference map.
This map assigns behavioral costs to each possible gaze location dependent on the last two fixations:
\begin{align}
r_\text{internal}(s,\mathbf{x_{t+1}}) &= r_\text{internal}(\mathbf{x_{t-1}}, \mathbf{x_t}, \mathbf{x_{t+1}}) \nonumber \\
&= \psi_0 \left(\Vert\mathbf{x_{t+1}}-\mathbf{x_{t}}\Vert \right) \nonumber \\
&+ \psi_1 \arccos\left(\frac{(\mathbf{x_{t+1}}-\mathbf{x_{t}})\cdot(\mathbf{x_{t}} - \mathbf{x_{t-1}})}{\Vert\mathbf{x_{t+1}}-\mathbf{x_{t}}\Vert \Vert\mathbf{x_{t}}-\mathbf{x_{t-1}}\Vert} \right) \nonumber \\
&+ \psi_2 \arccos \left(\frac{(\mathbf{x_{t+1}}-\mathbf{x_{t})}\cdot  [1 \quad 0]}{\Vert\mathbf{x_{t+1}}-\mathbf{x_{t}}\Vert } \right) 
\label{eq:cost}
\end{align}
Note that it is crucially important to estimate the human preferences for gaze shifts independently from image content instead of estimating probabilities of gaze shift parameters from free view data, because the observed gaze shifts depend both on image content and the preferences in such datasets. 
\begin{figure*}[b!]
\begin{center}
\includegraphics[width=1.0\textwidth]{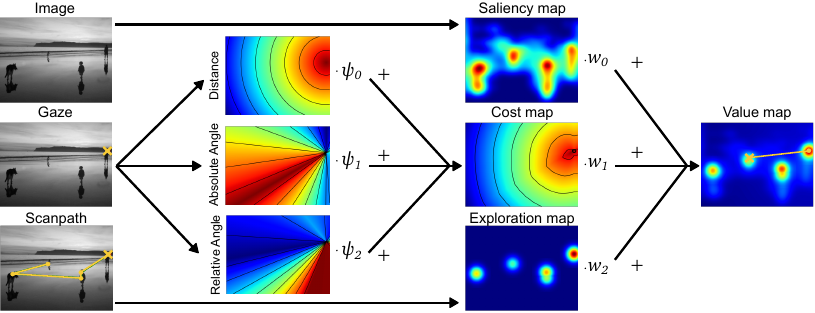}
\end{center}
   \caption{Schematic of the algorithm. An arbitrary saliency map and the scanpath with the current gaze position are the input. Output is a value map, which integrates the saliency map, the recomputed map for the cost of gaze shifts, and the sequential history dependant map. Note that the original image is not an input to the algorithm.}
\label{schematic}
\end{figure*}
\subsection{Approximating the value map}
We proposed three factors contributing to the final reward of an image location: task-related reward (Equation \ref{eq:rtask}; approximated through saliency), fixation history (Equation \ref{eq:fix_hist}) and the oculomotor costs (Equation \ref{eq:cost}).
To account for the sequential nature of visual scanpaths we extend static saliency approaches using an additional reward component, which is an exploration part based on past fixations. 
Note however, that only the reward component associated with the free viewing task is dependent on the image content whereas both the internal costs and the fixation history dependent part are independent of the image content. 

By consistently formulating the components as rewards we can combine them to yield the desired reward function:
\begin{align}
    r(s,\mathbf{x_{t+1}}) & = w_0 r_{\text{free view}}(s,\mathbf{x_{t+1}}) + w_1 r_{\text{internal}}(\mathbf{x_{t-1}}, \mathbf{x_{t}}, \mathbf{x_{t+1}}) \nonumber \\
    & + w_2 r_{\text{fixation history}}(\mathbf{x_{0}},\dots,\mathbf{x_{t-1}},\mathbf{x_t},\mathbf{x_{t+1}}) \nonumber \\
    & \approx w_0 S(I,\mathbf{x_{t+1}}) + w_1 \sum\limits_{i \in \{0,1,2\}} \psi_i (\mathbf{x_{t-1}}, \mathbf{x_t}, \mathbf{x_{t+1}}) \nonumber \\
    & + w_2 \sum_{i=0}^{t} \phi_i \mathcal{N}(\mathbf{x_{t+1}}; \mathbf{x_{i}},\mathbf{\Sigma}) 
\label{eq:qtotoal}
\end{align}
The parameters $w_0$, $w_1$, $w_2$ are linear weights and control the trade-off between task-related rewards, fixation history dependant rewards and internal costs and were estimated from the data.
We set $w_0$ equal to $1$, since the scale of our final value map does not matter and for the purpose of interpretability of the other parameters.

Computing the optimal policy in the MDP framework according to the reward function specified in Equation \ref{eq:qtotoal} would now require knowledge of the transition function, i.e. the state dependent gaze dynamics and their associated stochasticity.
Similarly, knowledge of sensory uncertainties would be needed across all possible tasks in order to find the optimal gaze shift policy within the POMDP framework.
Unfortunately, both these approaches are unfeasible. 
Instead, we use the common approximation of selecting the optimal one-step look-ahead action, i.e. greedy approximation by selecting the action that maximizes the reward for a single subsequent gaze shift. 

\begin{figure*}[t!]
\begin{center}
\includegraphics[width=0.9\textwidth]{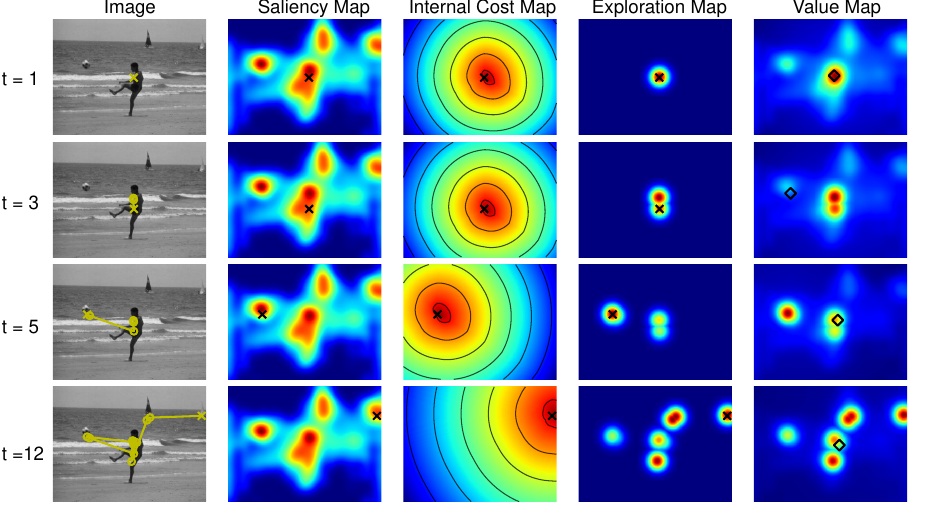}
\end{center}
   \caption{Example predictions of the next fixation. Each row shows the original image together with the respective preceding scanpath together with the current $i$-th fixation marked with a cross. The corresponding saliency, cost, and exploration maps as well as the final value map are shown from left to right. The  predicted fixation is shown together with the ground truth next fixation of the human observer marked with a diamond.}
\label{fig1}
\end{figure*}

The approximate value map depends on the image (through $S$), on the location of the last fixation (through the internal costs) and on the entire sequence of past fixations (through the history dependant part).
Crucially, as a consequence, the value map changes with every new fixation. 
The procedure of the computation of $Q$ is illustrated in Figure \ref{schematic} and examples of the respective maps for a succession of fixations is shown in Figure \ref{fig1}. 
\begin{algorithm}[!t]
\caption{Compute history dependent value map $V$ at timestep $t$}
\begin{algorithmic}
\STATE \textbf{Input:} Arbitrary saliency map $S$ from Image $I$, human scanpath $\mathbf{X} = \{\mathbf{x_0},\mathbf{x_1},...,\mathbf{x_t} \}$
\FORALL{possible fixations $\mathbf{x}$}
\STATE $C[\mathbf{x}] = \sum_{i \in \{0,1,2\}} \psi_i(\mathbf{x_{t-1}},\mathbf{x_t},\mathbf{x})$ 
\STATE $E[\mathbf{x}] = \sum_{i=0}^{t} \phi_i \mathcal{N}(\mathbf{x};\mathbf{x_{i}},\mathbf{\Sigma})$ 
\STATE $V[\mathbf{x}] = w_0 S[\mathbf{x}] + w_1 C[\mathbf{x}] + w_2 E[\mathbf{x}]$
\ENDFOR
\RETURN $V$
\end{algorithmic}
\label{algo:valuemap}
\end{algorithm}
Based on the approximate value map we can predict the future fixation locations from the policy $\pi$ based on the value map $Q(s,\mathbf{x_{t+1}})$, see Algorithm \ref{algo:valuemap}. 

\section{Experiments}
First, to demonstrate the utility of our algorithm in improving the prediction of the next fixation of human observers for arbitrary saliency models,
our model was implemented with four different underlying saliency models, which are currently among the ten best on the  MIT/Tuebingen saliency benchmark \cite{kummerer_mittubingen_nodate} with respect to several evaluation metrics: DeepGaze II \cite{kummerer_understanding_2017-1}, SAM-ResNet \cite{cornia_predicting_2018}, EML-NET \cite{EML_Net} and CASNet II \cite{CASNet}.

The parameters describing the three components of the behavioral costs for gaze switches
corresponding to internal motor and cognitive costs were recently estimated in a psychophysical experiment from eye movement data collected in a preference elicitation paradigm \cite{hoppe2022costs} \footnote{The estimated cost structure is available from \cite{hoppe2022costs}}.
We collected a total of 70643 gaze shifts across 14 subjects following the experimental paradigm described in \cite{hoppe2022costs}.
Values for the cost dimensions saccade amplitude, relative angle, and absolute angle were estimated using a random utility model \cite{train2009discrete}.
The utility function was computed as the weighted sum of the individual dimensions.

To include the exploration map and calculate the resulting value map, the corresponding free parameters had to be estimated. Since 
we want to evaluate the prediction of the next $n$ fixations,
we need to find a metric suitable for comparing 
individual fixations. 
We chose the Normalized Scanpath Saliency metric \cite{PETERS20052397}, which is definied as:
\begin{equation}
    \text{NSS}(S,\mathbf{x_0}, \dots , \mathbf{x_T}) = \frac{1}{T} \sum_{i=0}^{T} S_Z(\mathbf{x_{i}}).
\end{equation}
where $T$ is the total amount of fixations for the current image. Here $S_Z$ is the saliency map standardized by its mean $\mu_S$ and its standard deviation $\sigma_S$
\begin{equation}
    S_Z = \frac{S - \mu_S}{\sigma_S}
\end{equation}
Thus, the metric can be viewed as an average of the standardized saliency scores at the corresponding fixation locations.
For more details on the metric score see e.g. \cite{kummererSaliencyBenchmarkingMade2018,bylinskii_what_2019,Judd_2012}. Since this method does not compare two continuous maps, but also considers the actual set of fixations in addition to the saliency map \cite{le_meur_methods_2013}, the metric is also suitable for our case of one-step or $n$-step ahead prediction. In this case, we do not average over the entire gaze sequence, but optimize the value map of our model so that there is as much mass as possible at the location of the next fixation.

More specific, a random subset of 10000 real human fixations was selected and the parameters of our model were optimized so that the value map could predict the single subsequent fixation as well as possible and thus maximizes the NSS score. All optimizations were done on the MIT1003 dataset \cite{judd_learning_2009}.
All selected fixations were between the third and eleventh gaze target in their respective sequence. This fixation interval was chosen so that at least two fixations had already been carried out by the human observers to be able to compute the cost map and 
because only about one percent of all fixation sequences contain more than ten fixations.

We estimated the exploration values $\phi_i$, the covariance matrix $\mathbf{\Sigma}$ for the Exploration Map (Equation \ref{eq:fix_hist}) and the weight parameters $w_1$,$w_2$ (Equation \ref{eq:qtotoal}) through gradient based optimization.
Note that the covariance matrix $\mathbf{\Sigma}$ was constrained to be a multiple of the identity matrix $\sigma^2 \mathbf{I}$.
We fixed the weight for the saliency map to one, so that the estimated parameters $w_1$,$w_2$ can be interpreted as quantifying the relative contributions of the costs for gaze switches and the history dependent reward relative to the saliency value.
We used the Limited-memory BFGS-B algorithm \cite{LBFGS} given the NSS score as an objective function to be maximized.
The hyperparameters can be found in Section \ref{sec:params} in the Supplementary Material.
In addition, to meet the computational cost of the multidimensional problem, the images were reduced by a factor of ten in both dimensions using bilinear interpolation.

\begin{figure}[b!]
\begin{center}
\includegraphics[width=0.5\textwidth]{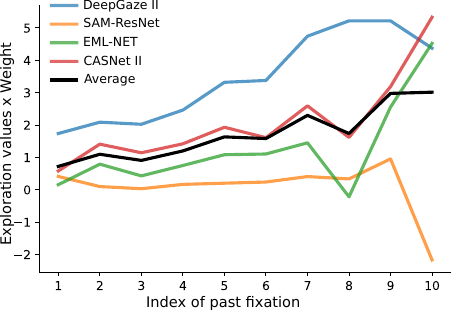}
\end{center}
   \caption{Estimated exploration values for four different underlying saliency models (blue, orange, red, green) and the corresponding averaged curve (black). Note that the estimated $\phi_i$ from Table S1 have been multiplied here by their associated weight $w_2$ to visualize the actual influence of the exploration map.}
\label{fig:exp_weights}
\end{figure}

Computations were performed on a high performance computer cluster.
All simulations were run on nodes with an Intel Xeon Processor E5-2680 v3 processor (2.5 GHz processor rate and 2.4 GB RAM).
The results of the optimization for all parameters can be found in the Supplementary Material in Table \ref{tab:param_estimation}.
Additionally, the estimated exploration values, and thus the weighting of past fixations are shown for all four models over time in Figure \ref{fig:exp_weights}.
These intermediate results were used to derive general weights of past fixations independent of the particular model saliency.
To this end, the estimated exploration values of the four models were averaged to be flexibly applied to arbitrary, new models. The resulting distribution is shown by the bold black curve in Figure \ref{fig:exp_weights}. 

The same experiment was repeated for all saliency models and image databases, except that the weight parameters were no longer co-estimated, i.e. the previously determined values were used. The results of this experiment can be found in Table S2 in the Supplementary Material. In addition, a new model was evaluated, which also belongs to the top evaluated saliency models on the MIT/Tuebingen benchmark - UNISAL \cite{drostejiao2020}.
This was to 
investigate the degree to which the optimized model parameters would
generalize from the four baseline models to a new model.

\section{Results}
We evaluated our method on three frequently used benchmarks, the MIT1003 \cite{judd_learning_2009}, the OSIE \cite{OSIE} and the Toronto dataset \cite{Toronto}. The MIT1003 and the OSIE dataset contain eye movements of 15 subjects during a three-second free viewing task on 1003 and 700 natural indoor and outdoor scenes, respectively. The Toronto dataset consits of 20 subjects during a four-second free viewing task on 120 color images of outdoor and indoor scenes.

\subsection{One-step ahead predictions}
We evaluated the one-step ahead predictions of our model with the NSS metric on the three datasets.
Additionally we used a second metric, the Area under Curve (AUC) (see \cite{Koenig2011,kummererSaliencyBenchmarkingMade2018,bylinskii_what_2019,Judd_2012} for details) for a second evaluation measurement, which was not considered during optimization.
AUC is also a well known hybrid measure for evaluating fixation prediction and saliency models \cite{le_meur_methods_2013}, which can be understood as a binary classifier for whether pixels are fixated or not.

These two metrics can be used in evaluating the prediction of the next fixation, see \cite{kummerer2021state}.
Other saliency metrics, like KL-divergence, Correlation Coefficient or Information gain are distribution based, so they assume the ground truth map to be a density and not a single fixation.
Therefore, they cannot be used to evaluate models predicting the next gaze target or any other per fixation evaluation. 
Example images with best and worst NSS scores are provided in Figure \ref{fig:example_best} and \ref{fig:example_worst} of the Supplementary Material.
Regarding scanpath prediction metrics (like ScanMatch or MultiMatch), we follow the evidence and argumentation from \cite{kummerer2021state}, arguing that it makes more sense to evaluate the capability of a model to predict the next fixation, which is exactly what saliency metrics do.

For evaluation, both metrics were calculated on all fixations of the three datasets above.
For the first fixation, the model selects a target exclusively based on the saliency map as neither the internal cost nor an fixation history can contribute. 
To predict the second fixation, we assumed that the fixation prior to image onset was at the image's center. This is true for most experiments and this only influences the relative angle of the cost map.

\begin{table*}[!htb]
    \caption{Evaluation results. AUC and NSS scores for the one-step and two-step ahead prediction of gaze targets based on sequential value maps compared to the respective saliency model's baseline}
    \label{tab:metrics}
    \begin{minipage}{.5\linewidth}
      \caption*{(a) One-step ahead predictions}
      \label{tab:metrics_one}
      \centering
      \begin{adjustbox}{max width=\textwidth}
        \begin{tabular}{m{2.5cm} cc|cc|ll}
              & \multicolumn{2}{c|}{MIT 1003}   & \multicolumn{2}{c|}{OSIE}       & \multicolumn{2}{c}{Toronto}                       \\
              & AUC            & NSS            & AUC            & NSS            & \multicolumn{1}{c}{AUC} & \multicolumn{1}{c}{NSS} \\ \hline
DeepGaze II   & 0.844          & 1.506          & 0.906          & 1.867          & 0.497                   & -0.031                  \\
Our extension & \textbf{0.874} & \textbf{1.856} & \textbf{0.908} & \textbf{2.569} & \textbf{0.632}          & \textbf{0.823}          \\ \hline
SAM-ResNet    & 0.864          & 2.222          & 0.905          & 3.088          & 0.477                   & -0.105                  \\
Our extension & \textbf{0.881} & \textbf{2.323} & \textbf{0.917} & \textbf{3.315} & \textbf{0.639}          & \textbf{0.706}          \\ \hline
EML-NET       & 0.864          & 2.255          & 0.902          & 3.050          & 0.490                   & -0.073                  \\
Our extension & \textbf{0.882} & \textbf{2.329} & \textbf{0.919} & \textbf{3.330} & \textbf{0.638}          & \textbf{0.656}          \\ \hline
CASNet II     & 0.860          & 1.993          & 0.898          & 2.587          & 0.515                   & -0.059                  \\
Our extension & \textbf{0.879} & \textbf{2.155} & \textbf{0.915} & \textbf{3.003} & \textbf{0.684}          & \textbf{1.033}          \\ \hline
UNISAL        & 0.889          & 2.612          & 0.890          & 2.755          & 0.542                   & 0.020                   \\
Our extension & \textbf{0.898} & \textbf{2.653} & \textbf{0.909} & \textbf{3.159} & \textbf{0.626}          & \textbf{0.451}         
\end{tabular}
\end{adjustbox}
    \end{minipage}%
    \begin{minipage}{.5\linewidth}
      \centering
        \caption*{(b) Two-step ahead predictions}
        \label{tab:metrics_two}
        \begin{adjustbox}{max width=0.75\textwidth}
        \begin{tabular}{cc|cc|ll}
              \multicolumn{2}{c|}{MIT 1003}   & \multicolumn{2}{c|}{OSIE}       & \multicolumn{2}{c}{Toronto}                       \\
              AUC            & NSS            & AUC            & NSS            & \multicolumn{1}{c}{AUC} & \multicolumn{1}{c}{NSS} \\ \hline
  0.844          & 1.506          & 0.906          & 1.867          & 0.497                   & -0.031                  \\
 \textbf{0.8554} & \textbf{1.725} & 0.888 & \textbf{1.899} & \textbf{0.602}         & \textbf{0.624}          \\ \hline
     0.864          & 2.222          & 0.905          & 3.088          & 0.477                   & -0.105                  \\
  \textbf{0.862} & \textbf{2.301} & 0.894 & 2.862 & \textbf{0.598}          & \textbf{0.535}          \\ \hline
        0.864          & 2.255          & 0.902          & 3.050          & 0.490                   & -0.073                  \\
  \textbf{0.869} & \textbf{2.332} & \textbf{0.903} & 2.897 & \textbf{0.601}          & \textbf{0.508}          \\ \hline
      0.860          & 1.993          & 0.898          & 2.587          & 0.515                   & -0.059                  \\
  \textbf{0.865} & \textbf{2.098} & 0.894 & 2.475 & \textbf{0.608}         & \textbf{0.598}          \\ \hline
         0.889          & 2.612          & 0.890          & 2.755          & 0.542                   & 0.020                   \\
 0.888 & \textbf{2.667} & \textbf{0.893} & \textbf{2.841} & \textbf{0.604}    & \textbf{0.371}        
\end{tabular}
\end{adjustbox}
\end{minipage} 
\end{table*}

The baseline saliency models were evaluated equivalently, but instead of using our dynamic value maps, the static history-independent maps were used. The results on the three different datasets with five different baseline models are shown in Table \ref{tab:metrics_one}(a). We reached higher scores on all three datasets compared to all baseline models, even for the UNISAL model, which was not used in the estimation of the parameters of the exploration map. These results transferred in all cases to the AUC score, which had not been used in the optimization. Thus, subsequent fixations on the datasets are better predicted by our dynamic one-step ahead prediction maps compared to the static baseline saliency models. This provides evidence, that including the independently measured human cost function for carrying out eye movements improves predictions by saliency maps.

\subsection{n-step ahead predictions}
Although the free parameters of the model were optimized to maximize predictions of the single next fixation on the NSS score for the MIT1003 data set, we can test the performance of the $n$-step predictions. Table \ref{tab:metrics_two}(b) and Table \ref{tab:supp_metrics} in the Supplementary Material report the results of the two-step and three-step predictions respectively. These results show, that the present model performs better consistently on the NSS score for both the MIT1003 and Toronto datasets across the second and third fixation predictions for all tested saliency models. Performance on the AUC score starts deteriorating for the prediction of the third fixation on the MIT1003 dataset but not the Toronto dataset. By comparison, both AUC and NSS scores are weaker already for the predictions of the second fixations on the OSIE dataset for all tested saliency models. 

\subsection{Influence of past fixations}
In addition to predicting the next fixation in a gaze sequence, our model allows quantifying and explaining the relative influence of past fixations. Since the exploration values $\phi_i$ were not constrained, we are able to interpret them directly. Figure \ref{fig:exp_weights} shows the relative value of past fixations over time. Overall, the value of refixating an image location increases approximately linearly over time. This indicates that having visited location $\mathbf{x_i}$ $i$ fixations ago during the same gaze sequence increases the probability of targeting the next fixation at location $\mathbf{x_i}$. This effect increases with increasing $i$, which means that fixation locations visited longer ago become more attractive for the observer.
\begin{figure*}[t]
\begin{center}
\includegraphics[width=\textwidth]{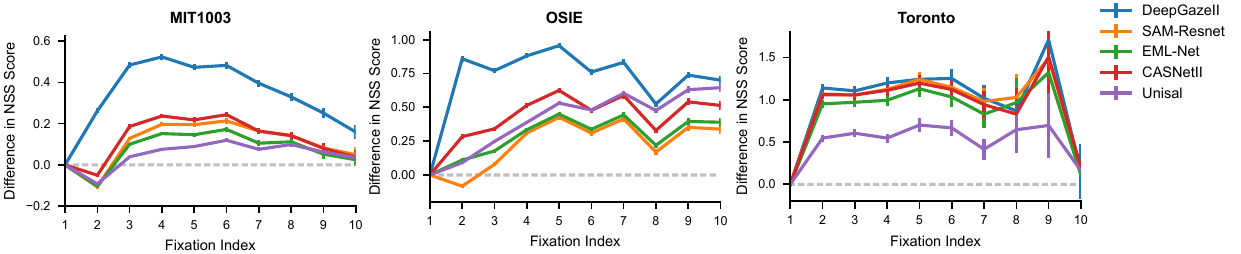}
\end{center}
   \caption{Differences in the NSS scores between our dynamic value maps and the underlying static saliency maps. Positive values indicate that our dynamic model predicted the subsequent fixation better than the baseline model. The errorbars indicate $\pm$ standard error of the mean.}
\label{fig:deltas}
\end{figure*}

For further analysis, we can quantify how well our predictions work for individual ordinal positions in the gaze sequence.
For this, we selected all predictions by their ordinal position and averaged the NSS scores grouped by their fixation index.
The progression of the goodness of the predictions can be seen in Figure \ref{fig:nss_by_pos} for all five models on all three datasets in comparison to the underlying baseline saliency models.
The differences in NSS scores can be seen in Figure \ref{fig:deltas}.
These results demonstrate, that the prediction accuracy is higher throughout the entire sequence up to the tenth gaze target, which was the last considered for almost all combinations of saliency models and image data sets.
This further supports the usefulness and validity of the current approach.

\section{Discussion}
In this paper, we introduced a computational model utilizing arbitrary saliency maps for computing sequential value maps to predict the next gaze target in human fixation sequences \cite{kummerer2021state}.
We conceptualized gaze sequences as sequential decision making within the framework of statistical decision theory, similar to previous approaches \cite{mathe2013action,jiang_learning_2016,hoppe2019multi}.
The model converts static saliency maps into a sequence of value maps, which incorporate saliency as a general conspicuity value across tasks, intrinsic human costs for gaze shifts, and a sequential history dependant reward. 
Given a saliency map of arbitrary origin and a sequence of previous gaze targets on an image,
the model generates  predictions of the next most likely fixation. 
The intrinsic preferences for gaze shifts  used in the algorithm were recently estimated through a preference elicitation experiment independently of image content \cite{hoppe2022costs}  
and the spatial and temporal parameters of the influence of fixation history were inferred based on the MIT1003 data set. 
Finally, the relative contributions of the three value maps were optimized on the same data set to maximize prediction of the next fixation.
The algorithm can be applied to arbitrary saliency models and is available upon request from the authors.

The results demonstrate that the three components of the intrinsic costs for human gaze shifts \cite{hoppe2022costs}
are sufficient to improve predictions of subsequent gaze targets obtained from a saliency model. 
These results are evidence that the common simplifying assumption that human scan paths are independent of behavioral preferences in gaze selection does not hold. Instead, the analysis of the distribution of preferred angles demonstrates, that image content and preferences in gaze shifts interact in non-trivial ways, a fact that has previously been demonstrated empirically \cite{foulsham2008turning}. 
Although some previous approaches in scanpath modeling have acknowledged or implemented statistics of human gaze shifts \cite{boccignone_modelling_2004,wang_simulating_2011,tavakoli2013stochastic,le2017age,zanca2019gravitational}, these were not measured independently of image content. The problem this gives rise to, is that the empirical statistics e.g. of saccade lengths measured in free viewing is the result of the preferences for gaze shifts and the distribution of image features.
Thus, predictions of the next fixation need to be generated by taking the actual human costs of gaze shifts into account instead of the empirical distributions of gaze obtained from the databases, because the latter are the result of the interaction between image features and the costs for gaze shifts. Further research will evaluate, how frequently applied heuristics including inhibition of return, center bias, and proximity preference arise from the interactions of an observer with a visual scene building on concepts derived in this work.

\section{Acknowledgements}
Calculations for this research were conducted on the Lichtenberg high performance computer of the TU Darmstadt. FK and CR acknowledge the support by the German Research Foundation (DFG, grant: RO 4337/3-1; 409170253). This research was supported by the Hessian research priority program LOEWE within the project WhiteBox and the cluster projects “The Adaptive Mind” and “The Third Wave of AI” as part of the Excellence Program of the Hessian Ministry of Higher Education, Science, Research and Art.

\bibliographystyle{splncs04}
\bibliography{egbib}
\newpage
\renewcommand{\figurename}{Supplementary Figure}
\renewcommand{\tablename}{Supplementary Table}
\renewcommand{\thefigure}{S\arabic{figure}}
\renewcommand{\thetable}{S\arabic{table}}
\renewcommand{\thesection}{S\arabic{section}}
\setcounter{figure}{0}
\setcounter{table}{0}
\setcounter{section}{0}
\setcounter{page}{1}
\pagenumbering{Roman}

\newgeometry{top=1in, bottom=1in, left=1in, right=1.25in}

{\huge Supplementary Information}

\section{Optimization of the model parameters} \label{sec:params}

\subsection{Hyperparameters for the Limited-memory BFGS-B algorithm}
All optimizations in this paper were done with the L-BFGS-B algorithm, and specifically it's \emph{scipy}\footnote{\url{https://docs.scipy.org/doc/scipy/reference/optimize.minimize-lbfgsb.html##optimize-minimize-lbfgsb}} implementation.
The following hyperparameters were used:
\begin{itemize}
\item Maximum number of variable metric corrections: $m_{\text{cor}} = 10 $
\item Tolerance limit for stopping criterion $\frac{f^k -f^{k+1}}{\max(|f^k|,|f^{k+1}|,1)} \leq f_\text{tol} \cdot \varepsilon$ where $\varepsilon$ is the machine precision:  $f_{\text{tol}} = 10^{-7}$
\item Tolerance limit for stopping criterion $\max{|\text{proj}(g_i) | i = 1, ..., n} <= p_\text{tol}$ with $\text{proj}(g_i)$ the $i$-th component of the projected gradient: $p_\text{tol} = 10^{-5}$
\item Gradient step size: $\epsilon = 10^{-8}$
\item Maximum number of function evaluations: $n_{\text{fun}} = 15000$
\item Maximum number of iterations: $n_{\text{iter}} = 15000$
\item Maximum number of line search steps per iteration: $n_{\text{ls}} = 20$ 
\end{itemize}

\subsection{Estimated parameters}
Below are the concrete values for all estimated parameters, for all models.
Table \ref{tab:param_estimation} shows the parameters for the experiments, where $\phi_i$ values where estimated for each model individually and Table \ref{tab:param_estimation2} for the experiments with fixed $\phi_i$ values.
The $\phi_i$ values for the second experiment where the weighted averages from the first one.
\begin{table}[h!]
\begin{center}
\begin{tabular}{m{1.9cm}  m{0.65cm} | m{0.65cm} |m{1cm} | >{\centering\arraybackslash}m{0.65cm} | >{\centering\arraybackslash}m{0.65cm} | >{\centering\arraybackslash}m{0.65cm} | >{\centering\arraybackslash}m{0.65cm} | >{\centering\arraybackslash}m{0.65cm} | >{\centering\arraybackslash}m{0.65cm} | >{\centering\arraybackslash}m{0.65cm} |>{\centering\arraybackslash}m{0.9cm} |>{\centering\arraybackslash}m{0.65cm} |>{\centering\arraybackslash}m{0.9cm}}
           & \multicolumn{1}{c|}{$w_1$} & \multicolumn{1}{c|}{$w_2$} & \multicolumn{1}{c|}{$\mathbf{\sigma}$} & \multicolumn{1}{c|}{$\phi_1$} & \multicolumn{1}{c|}{$\phi_2$} & \multicolumn{1}{c|}{$\phi_3$} & \multicolumn{1}{c|}{$\phi_4$} & \multicolumn{1}{c|}{$\phi_5$} & \multicolumn{1}{c|}{$\phi_6$} & \multicolumn{1}{c|}{$\phi_7$} & \multicolumn{1}{c|}{$\phi_8$} & \multicolumn{1}{c|}{$\phi_9$} & \multicolumn{1}{c}{$\phi_{10}$} \\ \hline
DeepGaze II & 0.345                    &  2.893                     & 34.158  & 1.737 & 2.087 & 2.022 & 2.462 & 3.319 & 3.376 & 4.744 & 5.219 & 5.218 & 4.374 \\ 
SAM-ResNet  & 0.007                    & 0.003                      & 93.337  & 0.410 & 0.097 & 0.031 & 0.165 & 0.201 & 0.237 & 0.407 & 0.333 & 0.952 & -2.17 \\
EML-NET     & 0.095                    & 0.481                      & 18.296  & 0.155 & 0.790 & 0.427 & 0.748 & 1.081 & 1.104 & 1.449 & -0.22 & 2.553 & 4.523 \\
CASNet II   & 0.157                    & 0.851                      & 22.328  & 0.580 & 1.408 & 1.142 & 1.419 & 1.930 & 1.608 & 2.592 & 1.616 & 3.185 & 5.331 \\
\end{tabular}
\end{center}
\caption{Estimated model parameters with individual exploration values.}
\label{tab:param_estimation}
\end{table}

\begin{table}[h!]
\begin{center}
\begin{tabular}{m{1.9cm}  m{0.65cm} | m{0.65cm} |m{1cm} | >{\centering\arraybackslash}m{0.65cm} | >{\centering\arraybackslash}m{0.65cm} | >{\centering\arraybackslash}m{0.65cm} | >{\centering\arraybackslash}m{0.65cm} | >{\centering\arraybackslash}m{0.65cm} | >{\centering\arraybackslash}m{0.65cm} | >{\centering\arraybackslash}m{0.65cm} |>{\centering\arraybackslash}m{0.9cm} |>{\centering\arraybackslash}m{0.65cm} |>{\centering\arraybackslash}m{0.9cm}}
           & \multicolumn{1}{c|}{$w_1$} & \multicolumn{1}{c|}{$w_2$} & \multicolumn{1}{c|}{$\mathbf{\sigma}$} & \multicolumn{1}{c|}{$\phi_1$} & \multicolumn{1}{c|}{$\phi_2$} & \multicolumn{1}{c|}{$\phi_3$} & \multicolumn{1}{c|}{$\phi_4$} & \multicolumn{1}{c|}{$\phi_5$} & \multicolumn{1}{c|}{$\phi_6$} & \multicolumn{1}{c|}{$\phi_7$} & \multicolumn{1}{c|}{$\phi_8$} & \multicolumn{1}{c|}{$\phi_9$} & \multicolumn{1}{c}{$\phi_{10}$} \\ \hline
DeepGaze II & 0.351                    & 1.989                    & 33.632   & $\vdots$ & $\vdots$ & $\vdots$ & $\vdots$ & $\vdots$ & $\vdots$ & $\vdots$ & $\vdots$ &  $\vdots$  & $\vdots$                  \\
SAM-ResNet  & 0.110                   & 0.510                   & 26.742  & $\vdots$ & $\vdots$ & $\vdots$ & $\vdots$ & $\vdots$ & $\vdots$ & $\vdots$ & $\vdots$ &  $\vdots$  & $\vdots$        \\
EML-NET     & 0.095                     &  0.619                 & 21.553    & 0.720 & 1.095 & 0.906 & 1.198 & 1.633 & 1.581 & 2.298 & 1.737 & 2.977 & 3.014 \\
CASNet II   &  0.160                   &   1.134                    & 25.961   & $\vdots$ & $\vdots$ & $\vdots$ & $\vdots$ & $\vdots$ & $\vdots$ & $\vdots$ & $\vdots$ &  $\vdots$  & $\vdots$               \\ 
UNISAL   & 0.061                    &  0.483                  & 12.643  & $\vdots$ & $\vdots$ & $\vdots$ & $\vdots$ & $\vdots$ & $\vdots$ & $\vdots$ & $\vdots$ &  $\vdots$  & $\vdots$              \\   
\end{tabular}
\end{center}
\caption{Estimated model parameters with fixed exploration values.}
\label{tab:param_estimation2}
\end{table}

\newpage
\section{Quality of the predictions depending on the ordinal position in the gaze sequence}
Here we show a detailed comparison between all baseline models and our one-step ahead extension on all three datasets, based on the ordinal position in the scanpath.
Only the first 10 fixations were included because for positions larger than 10, the number of fixations was very low.
The first position is always equal because our model uses only saliency information when there is no prior fixation information available.
\begin{figure}[h!]
    \centering
    \includegraphics[width=1\textwidth]{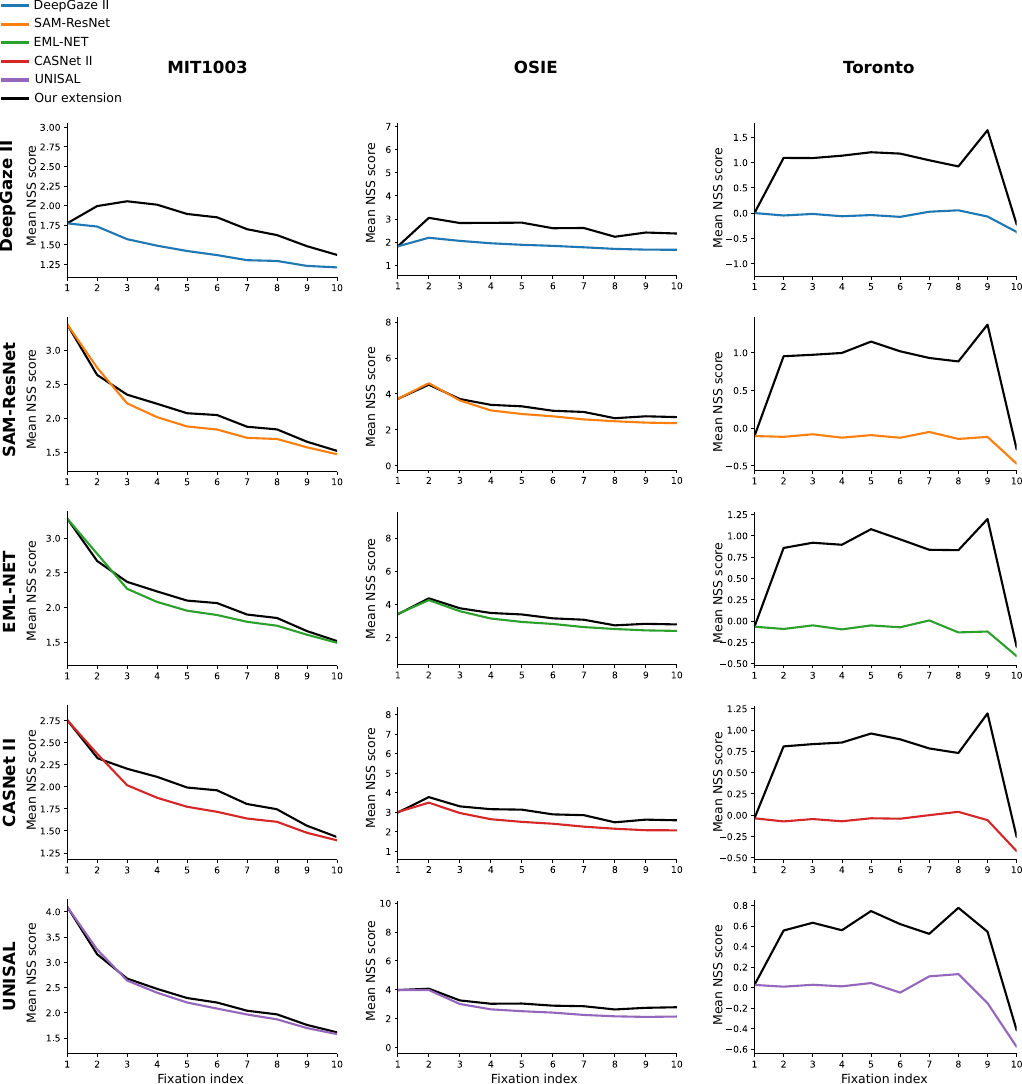}
    \caption{NSS scores for the one-step ahead prediction depending on the ordinal position in the gaze sequence.}
    \label{fig:nss_by_pos}
\end{figure}

\newpage
\section{Example predictions with best and worst NSS}
4 of the best and worst single fixations, with respect to NSS score of the OSIE/MIT 1003 dataset.
Shown is the image, with the scanpath up to that point (left), all three internal parts of our model (middle), and the resulting value map with the next fixation highlighted (right).
\begin{figure}[h!]
    \centering
    \includegraphics[width=0.8\textwidth]{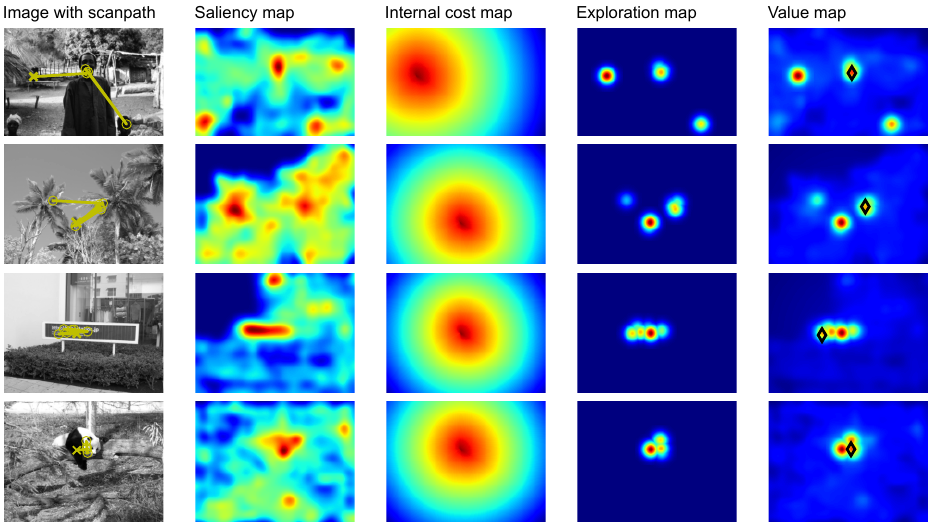}
    \caption{Examples of predictions of the next fixations with highest NSS score.}
    \label{fig:example_best}
\end{figure}
\begin{figure}[h!]
    \centering
    \includegraphics[width=0.8\textwidth]{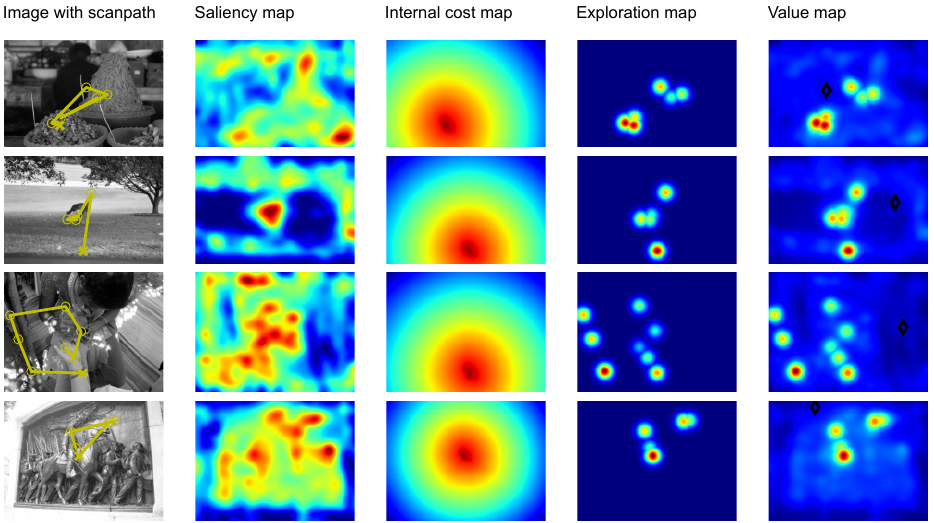}
    \caption{Examples of predictions of the next fixations with lowest NSS score.}
    \label{fig:example_worst}
\end{figure}

\newpage
\section{Three-step ahead predictions}
Similar to Table \ref{tab:metrics}, we evaluated also the three-step ahead predictions of the model, using only information up to timestep $t-3$.
Now our model is only marginally better than the respective baseline and is even worse on some datasets/baseline model combinations.
\begin{table*}[h!]
\begin{center}
{\small
\begin{tabular}{lcc|cc|ll}
              & \multicolumn{2}{c|}{MIT 1003}   & \multicolumn{2}{c|}{OSIE}       & \multicolumn{2}{c}{Toronto}                       \\
              & AUC            & NSS            & AUC            & NSS            & \multicolumn{1}{c}{AUC} & \multicolumn{1}{c}{NSS} \\ \hline
DeepGaze II   & 0.844          & 1.506          & 0.906          & 1.867          & 0.497                   & -0.031                  \\
Our extension & \textbf{0.850} & \textbf{1.662} & 0.885 & 1.784 & \textbf{0.565}          & \textbf{0.394}          \\ \hline
SAM-ResNet    & 0.864          & 2.222          & 0.905          & 3.088          & 0.477                   & -0.105                  \\
Our extension & 0.857 & \textbf{2.271} & 0.886 & 2.813 &    \textbf{0.557}      & \textbf{0.317}          \\ \hline
EML-NET       & 0.864          & 2.255          & 0.902          & 3.050          & 0.490                   & -0.073                  \\
Our extension & 0.864 & \textbf{2.304} & 0.867 & 2.851 & \textbf{0.564}          & \textbf{0.307}          \\ \hline
CASNet II     & 0.860          & 1.993          & 0.898          & 2.587          & 0.515                   & -0.059                  \\
Our extension & \textbf{0.861} & \textbf{2.062} & 0.888 & 2.406 & \textbf{0.576}          & \textbf{0.354}          \\ \hline
UNISAL        & 0.889          & 2.612          & 0.890          & 2.755          & 0.542                   & 0.020                   \\
Our extension & 0.885 & \textbf{2.646} & 0.887 & \textbf{2.786} & \textbf{0.582}         & \textbf{0.243}         
\end{tabular} }
\end{center}
\caption{Evaluation results. AUC and NSS scores for the three-step ahead prediction of gaze targets based on sequential value maps compared to the respective saliency model's baseline.}
\label{tab:supp_metrics}
\end{table*}

%
%
\end{document}